\documentclass[11pt,a4paper]{article}
\usepackage[hyperref]{acl2020}
\usepackage{times}
\usepackage{latexsym}
\usepackage{booktabs}
\usepackage{adjustbox}
\usepackage{multirow}
\usepackage{color, colortbl}
\definecolor{Gray}{gray}{0.93}

\newcommand{\shortname}{\textsc{Raddle}}
\newcommand{\gptft}{GPT-2$_{\texttt{FT}}$}

\newcommand{\iconvar}{\raisebox{-2pt}{\includegraphics[height=.4cm,width=0.4cm]{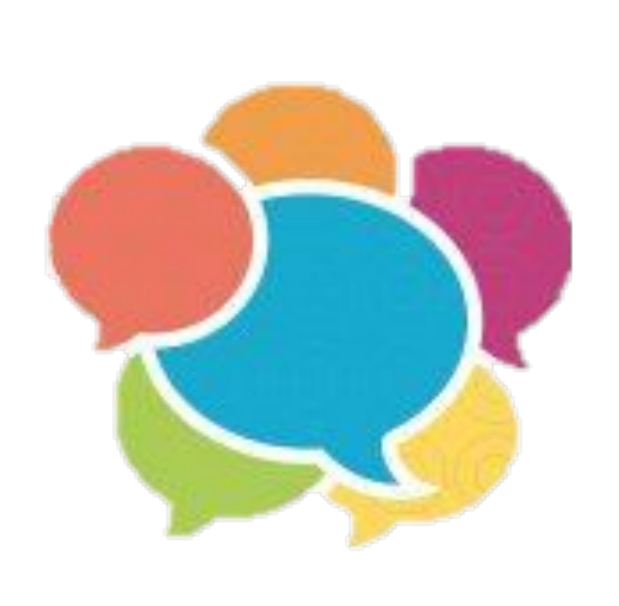}  }}
\newcommand{\iconasr}{\raisebox{-2pt}{\includegraphics[height=.4cm,width=0.4cm]{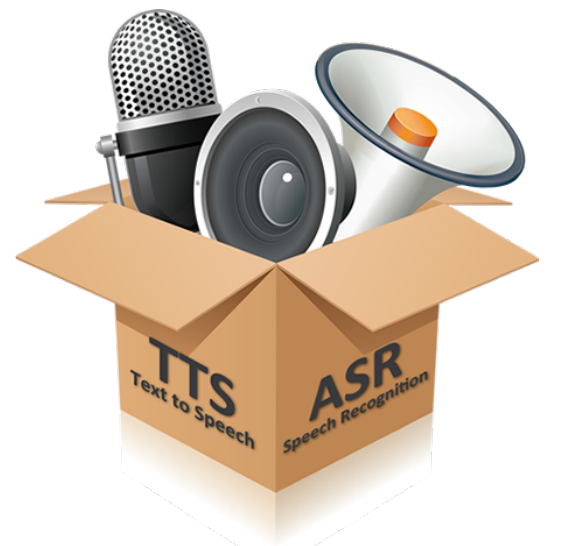}  }}
\newcommand{\iconent}{\raisebox{-2pt}{\includegraphics[height=.4cm,width=0.4cm]{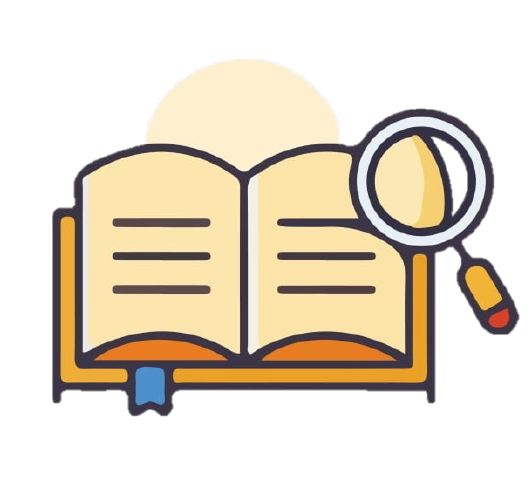}  }}
\newcommand{\iconood}{\raisebox{-2pt}{\includegraphics[height=.4cm,width=0.4cm]{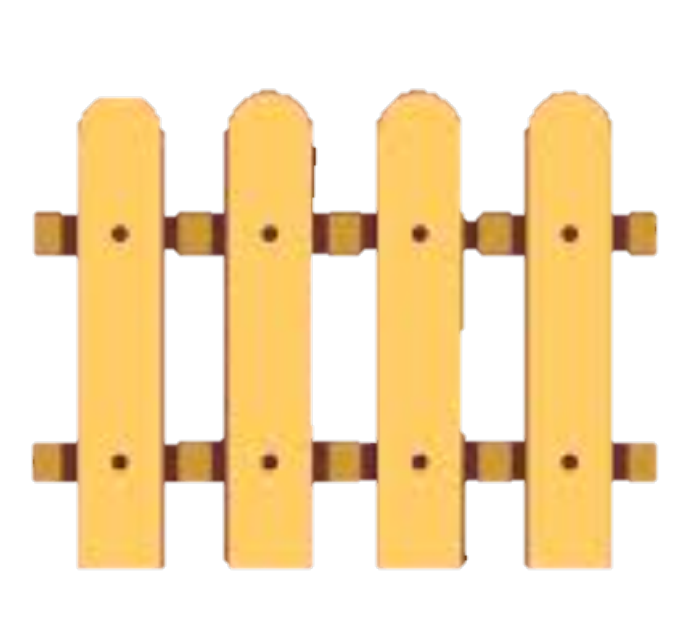}  }}

\makeatletter
\newcommand{\thickhline}{%
    \noalign {\ifnum 0=`}\fi \hrule height 1pt
    \futurelet \reserved@a \@xhline
}
\makeatother

    \makeatletter
\def\@fnsymbol#1{\ensuremath{\ifcase#1\or \dagger\or \ddagger\or
   \mathsection\or \mathparagraph\or \|\or **\or \dagger\dagger
   \or \ddagger\ddagger \else\@ctrerr\fi}}
    \makeatother

\usepackage{microtype}

\usepackage{amsmath}
\usepackage{amsfonts}
\usepackage{amssymb}
\usepackage{wrapfig}
\usepackage{subcaption}
\usepackage{multirow}
 \usepackage{mathtools} 

\usepackage{verbatim}

\usepackage{anyfontsize}

\usepackage{color}
\usepackage{tikz}
\usetikzlibrary{arrows,shapes,snakes,automata,backgrounds,fit,petri}
\usepackage{adjustbox}

\newcommand{\RN}[1]{%
	\textup{\lowercase\expandafter{\it \romannumeral#1}}%
}

\usepackage{booktabs}

\usepackage{tcolorbox}

\makeatletter
\newcommand{\distas}[1]{\mathbin{\overset{#1}{\kern\z@\sim}}}%

\usepackage{enumitem}

\usepackage[lined,boxed,commentsnumbered,ruled,linesnumbered]{algorithm2e}

\newcommand{\ie}[0]{\emph{i.e., }}

\newcommand{\etc}[0]{\emph{etc.}}

\newcommand{\beq}{\vspace{0mm}\begin{equation}}
\newcommand{\eeq}{\vspace{0mm}\end{equation}}
\newcommand{\beqs}{\vspace{0mm}\begin{eqnarray}}
\newcommand{\eeqs}{\vspace{0mm}\end{eqnarray}}
\newcommand{\barr}{\begin{array}}
\newcommand{\earr}{\end{array}}

\ifx\assumption\undefined
\newtheorem{assumption}{Assumption}
\fi

\ifx\definition\undefined
\newtheorem{definition}{Definition}
\fi

\ifx\remark\undefined
\newtheorem{remark}{Remark}
\fi

\aclfinalcopy 

\title{\shortname{}: An Evaluation Benchmark and Analysis Platform for\\ Robust Task-oriented Dialog Systems}

\author{Baolin Peng$^1$, Chunyuan Li$^1$, Zhu Zhang$^{12}$\thanks{~~Work was done when Zhu Zhang was visiting MSR} ~, Chenguang Zhu$^1$, Jinchao Li$^1$, Jianfeng Gao$^1$ \\
  $^1$Microsoft Research, Redmond, WA \\
  $^2$Iowa State University / Ames, IA \\
  \texttt{\{bapeng,chunyl,chezhu,jincli,jfgao\}@microsoft.com} \\
  \texttt{zhuzhang@iastate.edu}
 }

\date{}

\begin{document}
\maketitle
\begin{abstract}
For task-oriented dialog systems to be maximally useful, it
must be able to process conversations in a way that is (1) generalizable with a small number of training examples for new task domains, and (2) robust to user input in various styles, modalities or domains. In pursuit of these goals, we introduce the  \shortname{}\footnote{{\bf R}obust  t{\bf A}sk-oriente{\bf D} {\bf D}ia{\bf L}og systems {\bf E}valuation} benchmark~\footnote{Benchmark link: \url{http://aka.ms/raddle}}, a collection of corpora and tools for evaluating the performance of models across a diverse set of domains. By including tasks with limited training data, \shortname{} is designed to favor and encourage models with a strong generalization ability. \shortname{} also includes a diagnostic checklist that facilitates detailed robustness analysis in aspects such as language variations, speech errors, unseen entities, and out-of-domain utterances.
We evaluate recent state-of-the-art systems based on pre-training and fine-tuning, and find that grounded pre-training on heterogeneous dialog corpora performs better than training a separate model per domain. Overall,  existing models are less than satisfactory in robustness evaluation, which suggests  opportunities for future improvement.
\end{abstract}

\section{Introduction}

Dialogs constitute a crucial communication channel in completing a broad range of tasks, such as weather query, flight and restaurant booking, movie booking, IT helpdesk, \etc~
Comparing to chit-chat systems that are usually modeled with single-turn context-response pairs, task-oriented dialog systems involve retrieving information from knowledge bases and reasoning over multiple dialog turns. This makes it especially important for a system to be able to produce response that are grounded on tasks goals and user intents.
In a bid to support human-computer interactions, task-oriented dialog systems have been built to allow users to converse with a computer system using natural language, such as 
Siri~\footnote{\scriptsize \url{https://www.apple.com/siri/}}, Google Assistant~\footnote{\scriptsize \url{https://assistant.google.com/}}, Amazon Alexa~\footnote{\scriptsize \url{https://developer.amazon.com/en-US/alexa}}, Microsoft XiaoIce \cite{zhou2020design}. 
Traditionally, a task-oriented dialog system uses a modularized pipeline with four modules that execute sequentially \cite{gao2019neural}.
A natural language understanding ({\bf NLU}) module identifies user intents and extracts associated information such as slots and corresponding values from user input. A dialog state tracker ({\bf DST}) infers the belief state (or user goal) from dialog history.
The belief state is often used to query a task-specific database (DB) to obtain the DB state, such as the number of entities that match the user goal.
The dialog state and DB state are then passed to a dialog policy ({\bf POL}) module to select the next system action.
A natural language generation ({\bf NLG}) module converts the action to a natural language response.

The human ability to converse is general, flexible, and robust. In contrast, most popular tools for dialog system development adopting the above modular systems are designed for specific tasks and struggle with out-of-scope data. If we aspire to develop models beyond extensively hand-crafted rules and annotated data for each single domain/task, it is critical to develop a more {\em unified}, {\em efficient} and {\em robust} model that can more quickly learn to execute a range of different tasks in different domains.

To fuel research in this direction, we present the \shortname{} benchmark. It includes a collection of task-oriented dialog tasks in diverse domains (e.g. end-to-end modeling, dialog state tracking). The benchmark also has a companion online platform for model evaluation, comparison, and robustness analysis. Importantly, \shortname{} exhibits two unique advantages that pave the way for building  more pragmatic dialog systems:
$(\RN{1})$  {\bf Limited data setting} is the major focus of \shortname{}, to evaluate the generalization ability of models. It aims at simulating the real-world application scenarios where only very limited amount of labelled data is available for new domains. Given this focus, \shortname{} is therefore a favorable benchmark to evaluate recent models in the pre-training and fine-tuning paradigm, which learn to represent linguistic knowledge in a way that facilitates sample-efficient learning and effective knowledge transfer. 
$(\RN{2})$  {\bf Robustness analysis} is introduced to study model performance in various challenging scenarios, where models are evaluated with anomalous user input such as language variations,  speech  errors,  unseen  entities and  out-of-domain utterances. Failing to handle these inputs often produce inappropriate responses leading to frustrating user experience. These scenarios are common for deployed systems in the real world, but are largely ignored in existing dialog benchmarks.  To the best of our knowledge, \shortname{} presents the first work to fill this gap. 


To better understand the challenges posed by \shortname{}, we conduct experiments with simple baselines
and state-of-the-art task-oriented dialog models. We find that grounded pre-trained models with a unified multi-task learning objective outperform models separately trained on each domain. Moreover, even the best performing model (\textsc{Soloist}~\cite{peng2020soloist}) in our evaluation achieves a fairly low  score in robustness analysis. This suggests that our baseline models can handle common inputs with strong regularities, but struggle with anomalous inputs that require deeper reasoning.

In summary, our key contributions are:
$(\RN{1})$ A novel dialog benchmark with an emphasis on limited data and multiple domains/tasks, which formally creates a scenario to evaluate the grounding and generalization ability of pre-trained models. 
$(\RN{2})$ A crowd-sourced diagnostic evaluation dataset to cover a broad range of real-world sophistication to study model robustness.
$(\RN{3})$  An online evaluation platform and leaderboard to track research progress, with human evaluation services to be granted to top-ranked submissions on a bi-monthly basis.
$(\RN{4})$ Baseline results for major existing approaches to task-oriented dialogs.

\section{Related Work}
\subsection{Dialog Benchmarks} 

To drive the progress of building dialogue systems using data-driven approaches, a number of conversational corpora have been released. They are roughly grouped into two categories: 
$(\RN{1})$ Corpora with structured semantic labels~\cite{wen2016network,shah2018building}. These datasets are often specifically annotated, and used to study an individual module in the dialog pipeline. For example, DialoGLUE~\cite{mehri2020dialoglue} is a recently proposed benchmark with a focus on NLU and DST tasks. 
$(\RN{2})$ Corpora with an implicit user goal~\cite{lowe2015ubuntu}. These datasets are often without semantic labels but can be used in end-to-end (E2E) dialog modeling~\cite{li2016diversity,zhu2020boosting,wu2019alternating,zhu2019multi}. For example, ConvLab~\cite{convlab,convlab-2} is a recent platform for multi-domain E2E evaluation.

MultiWOZ~\cite{budzianowski2018multiwoz} is the most related work to \shortname{}. It is a large-scale
multi-turn conversational corpus across several domains. It can be
used to develop individual dialog modules as separate tasks for existing modular-based methods, or serves as a benchmark for E2E  dialogue modeling methods. \shortname{} inherits the advantages of MultiWOZ in its flexibility for separate/joint task modeling and its comprehensiveness in multi-domain data coverage, but differs significantly in two aspects: an emphasis on limited data settings and an unique robustness checklist. Both are essential qualities in building task bots at scale. 

Further, \shortname{} provides an online platform for model evaluation and fair comparison based on privately-held test data, inspired by GLUE~\cite{wang2018glue}. To the best of our knowledge, \shortname{} is the first online platform for DST and E2E tasks in the dialog community. This can reduce the inconsistency caused by different researchers/teams using varying processing/evaluation scripts to dilute where the gain comes from.


\begin{table*}[]
\centering
\small
\setlength{\tabcolsep}{1.2mm}{
\begin{tabular}{@{}lcccccccccc@{}}
\toprule
 & \multicolumn{4}{c}{Standard}                              & \multicolumn{4}{c}{Language variations / Speech Errors}  & Unseen   & OOD        \\ 
\cmidrule(l){2-5} \cmidrule(l){6-9} \cmidrule(l){10-10}  \cmidrule(l){11-11}  

Domain                        & {Attraction}   & {Train}        &  {Hotel}        & { Restaurant}   & { Attraction}   & {Train}        & {Hotel}        & {Restaurant}   & {Reminder} & {Attraction} \\ \midrule
\#Train                                     & 50           & 50           & 50           & 50           & -            & -            & -            & -            & 50       & 50         \\
\#Test                                      & 100          & 200          & 200          & 200          & 100          & 200          & 200          & 200          & 400      & 800        \\  \cmidrule(l){2-9}
Task                                        & \multicolumn{8}{c}{Dialog State Tracking / End-to-End Modeling} & DST / IC      & DST / OOD  \\  \cmidrule(l){2-9}
Metrics                                     & \multicolumn{8}{c}{Joint Goal Accuracy / Combined Score} & JGA / Acc.   & JGA / F1 \\
\bottomrule
\end{tabular}
}

\caption{Dataset descriptions and statistics. DST is short for Dialog State Tracking, E2E denotes End-to-end modeling, and $\mathtt{IC}$ stands for Intent Classification. Joint Goal Accuracy ($\mathtt{JGA}$) is used for DST and Combined score is used for E2E. }
    \label{table:statistics}
    
\end{table*}

\subsection{Evaluation of Pre-trained Models}

Pre-trained language models (PLMs) have substantially advanced the state of the art across a variety of language understanding and generation tasks~\cite{peters2018deep,devlin2019bert,yang2019xlnet,liu2019roberta,gpt2,keskar2019ctrl,dong2019unified,peng2020few,peng2020data,li2020optimus}.  PLMs are often trained to predict words based on their context on massive text data, and the learned models can be fine-tuned to quickly adapt to various downstream tasks, exhibiting strong generalization capacity even with just a few in-domain training examples. Building task bots at scale requires the model to deal with the limited data problem for each domain, which can be used as a testbed to evaluate the generalization ability of PLMs. To this end, we limit the number of task-specific training examples in \shortname{} to evaluate the sample-efficiency of models.

Meanwhile, task-oriented dialogues pose a unique set of challenges for PLMs \cite{gao2020robust}: a dialog is intrinsically goal-driven, multi-turn and often informal/noisy. Indeed, dialog-specific PLMs are proposed~\cite{wu2020tod,peng2020soloist}. 
However, the robustness of PLMs to linguistic perturbations often occurring in dialog settings (See Section 4 for details) is largely unexplored. Note that our notion of robustness emphasizes {\em natural} language variations, which is different from {\em adversarial} examples/training that aim to fool a trained model~\cite{nie2019adversarial}. From this perspective, \shortname{} provides an unique benchmark for assessing PLMs with a robustness orientation.

\section{Tasks}

\shortname{} is centered on five English dialog scenarios in daily life, which cover a broad range of data collection schemes, task types and complexities. As our first goal of \shortname{} is to spur development of generalizable
dialog systems, we design the benchmark such that a good performance requires a model to leverage
substantial knowledge (e.g., pre-trained parameters) learned from its previous life cycle, while still maintaining some task-specific components~\cite{DBLP:conf/acl/CoopeFGVH20, DBLP:conf/emnlp/HendersonCMSWV20,peng2020soloist,Wu2020ToDBERTPN}.
Specifically, we deliberately keep a small number of training examples for each scenarios. This is consistent with the common practice that only limited labelled data is provided when deploying a dialog system to new domains. Table \ref{table:statistics} shows the data statistics. Four domains in the standard-setting are sampled from MultiWOZ2.0~\cite{budzianowski2018multiwoz}. \texttt{Reminder} is intentionally only utilized for unseen entity tracking. Because it is a human-machine corpus with a relatively smaller action space meaning that the impact of policy learning on models is largely alleviated. Therefore, the performance of models on this corpus will mostly reflect its capability of unseen entity tracking.  Note that  The number of training examples is limited to 50, an accepted scale that users can provide.
Though it is possible to train a single model for each task from scratch without outside sources of knowledge, 
we expect that our focus on data-scarce settings will render this approach uncompetitive. 

Furthermore, a typical task-oriented dialog system uses a modularized pipeline that has four modules and executes sequentially. Recent research has shown promising results on parameterizing the modularized pipeline using a single neural auto-regressive model, and training it in an end-to-end manner \cite{peng2020soloist,Ham2020e2e,hosseini2020simple}. 
In fact, a single auto-regressive model can significantly ease the workflow of training and deploying dialog systems for new tasks, compared to existing modularized tools and methods. Therefore, we design the benchmark to allow evaluations on end-to-end dialog modeling, in addition to the modularized evaluation on dialog state tracking. To reveal the gap between the complexity of dialogues in lab environments and that in real scenarios, we construct a suite of tasks to study the robustness of models.  We describe these tasks below and in Table \ref{table:statistics}.

On the evaluation front, we concentrate on  simulation-based methodologies, in order to facilitate automation. Although we only  offer  human-based evaluations~\cite{gao2019neural} to top-ranked submissions at this point, we emphasize realistic scenarios in pursuit of system robustness (see Section~\ref{sec:robustness}).

\paragraph{Task 1: Dialog State Tracking}
A robust NLU and DST is the first step towards building a reliable dialog system.
The dialogue state is a summary of the entire conversation till the current turn. In a task-oriented
system, it is represented in the form of {\em slot-value} pairs, where {\em slot} indicates the category/attribute of the user goal expressed in the utterance, and {\em value} is the corresponding information.  For the evaluation metric, we report {\em joint goal accuracy}, which indicates the proportion of dialogue turns where all the user’s search goal constraints are correctly identified~\cite{mrksic2017neural}. 
To specially study the NLU performance, we consider {\em intent classification}, which aims to automatically extract meaning from a natural language utterance in order to understand user's goal~\cite{hemphill-etal-1990-atis,zhu2019sim}.

\paragraph{Task 2: End-to-end Modeling}
The end-to-end (E2E) dialog models consider dialog history as input, and produce the natural language response. It jointly implements the dialogue management (including DST and POL) and response generation (\ie NLG) components.
Following \citet{budzianowski2018multiwoz}, $\mathtt{Inform}$, $\mathtt{Success}$, and $\mathtt{BLEU}$ scores are reported.  The first two metrics evaluate dialog task completion:
$\mathtt{Inform}$ measures if the system provides a correct entity (inform rate), meanwhile $\mathtt{Success}$ measures the exact matching of answering all the requested information (success rate,) and if the answered information matches users' goal.
$\mathtt{BLEU}$ evaluates how fluent the generated responses are compared to human-written responses.
A combined score ($\mathtt{Combined}$) is also reported using $\mathtt{Combined} = (\mathtt{Inform} + \mathtt{Success}) \times 0.5 + \mathtt{BLEU}$ as an overall quality measure, as suggested in~\cite{budzianowski2018multiwoz}.

\section{Robustness Diagnostic Checklist} \label{sec:robustness}

Existing benchmarks assume a world of a ``perfect'' user who always provides precise, concise, and semantically unambiguous utterances. These goal-oriented dialog datasets are largely collected by crowd-sourcing, where a crowd-sourced worker enacts the part of a real user by following a set of template instructions provided for the task. This method results in a dataset where most user utterances are straight-forward, stick to the goal and tend to leave out the variation/errors commonly found in real-world conversational data. To this end, we collect a suite of language variations to reveal the dialog sophistication in the real world, and measure the robustness of dialog models.

\begin{figure*}[t!]
  \vspace{-0mm}\centering
  \begin{tabular}{c c}
    \hspace{-4mm}
    \includegraphics[height=3.2cm,width=7.80cm]{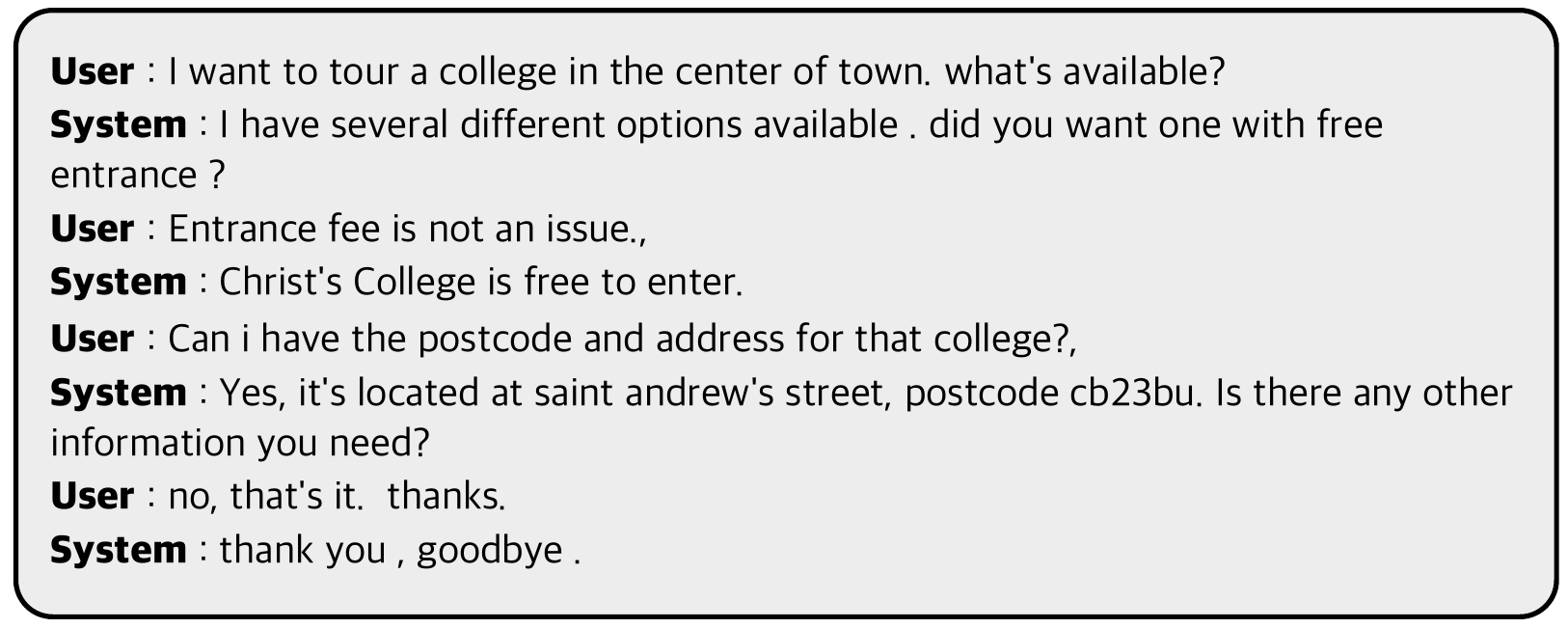} & 
    \hspace{-3mm}
    \includegraphics[height=3.2cm,width=7.80cm]{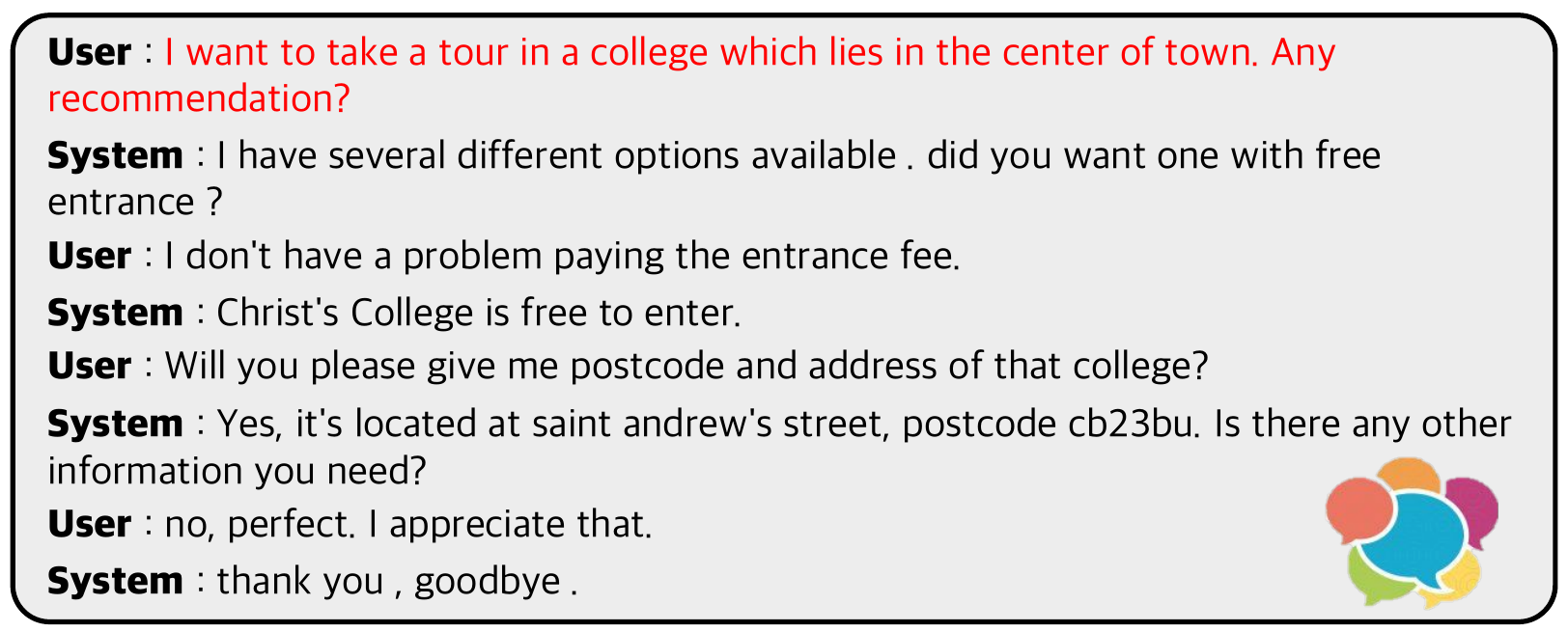} \\
    \vspace{3mm}
    \hspace{-6mm}
    (a) Standard dialog session \hspace{-3mm} & 
    (b) Paraphrase  \\ 
    \hspace{-4mm}
    \includegraphics[height=3.12cm,width=7.80cm]{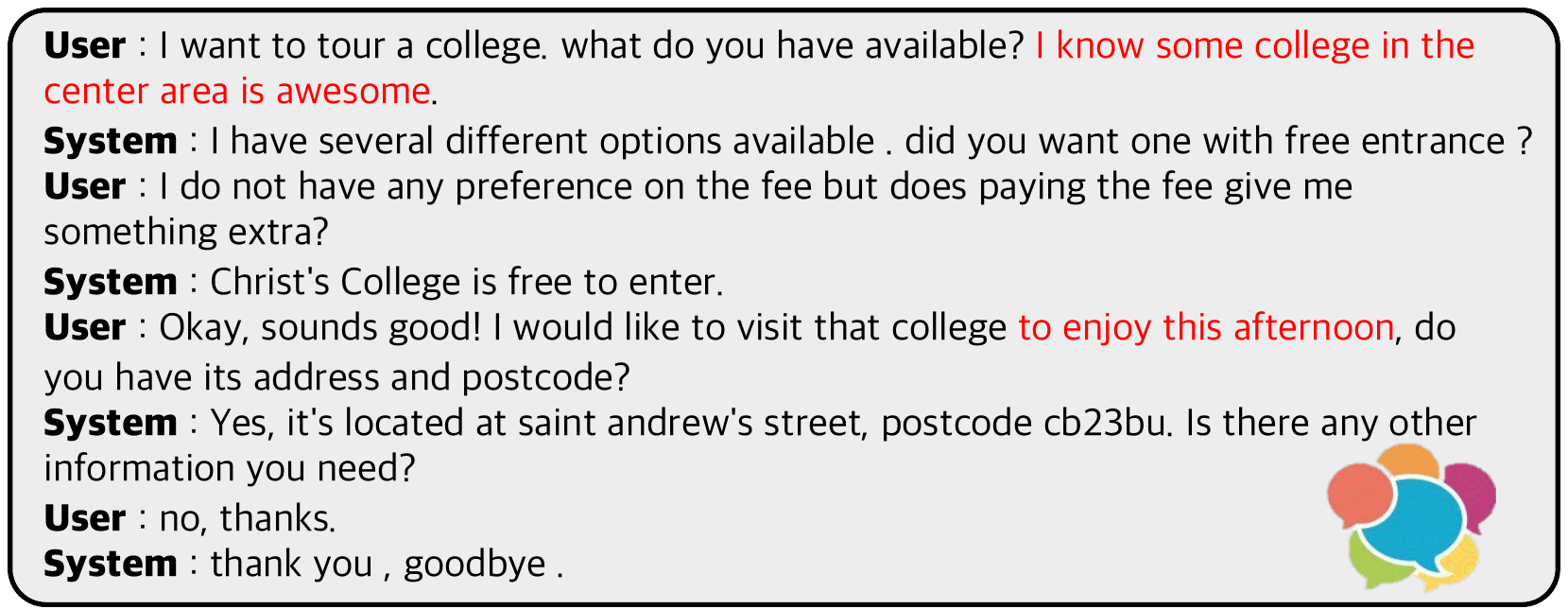} & 
    \hspace{-3mm}
    \includegraphics[height=3.15cm,width=7.80cm]{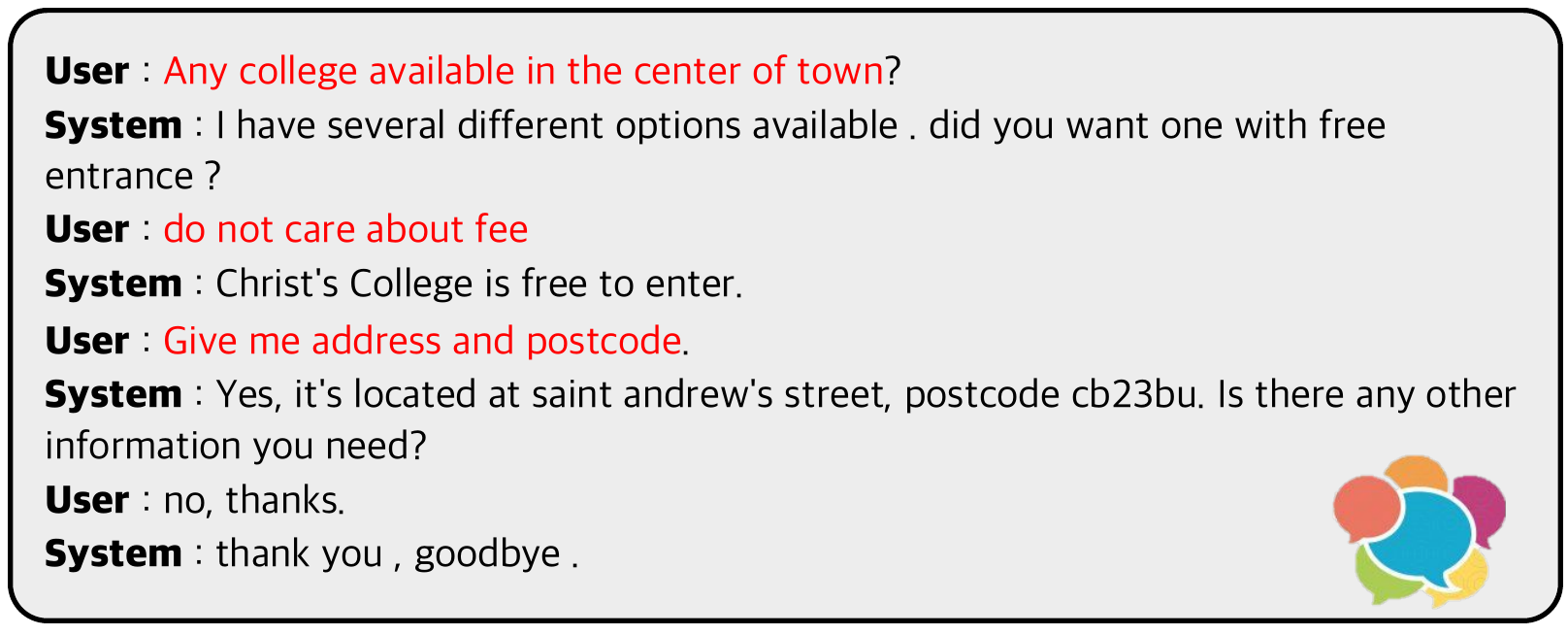} \\
    \vspace{3mm}
    (c) Verbosity \vspace{0mm} & 
    (d) Simplification \hspace{-0mm} \\ 
    \hspace{-4mm}
    \includegraphics[height=3.10cm,width=7.80cm]{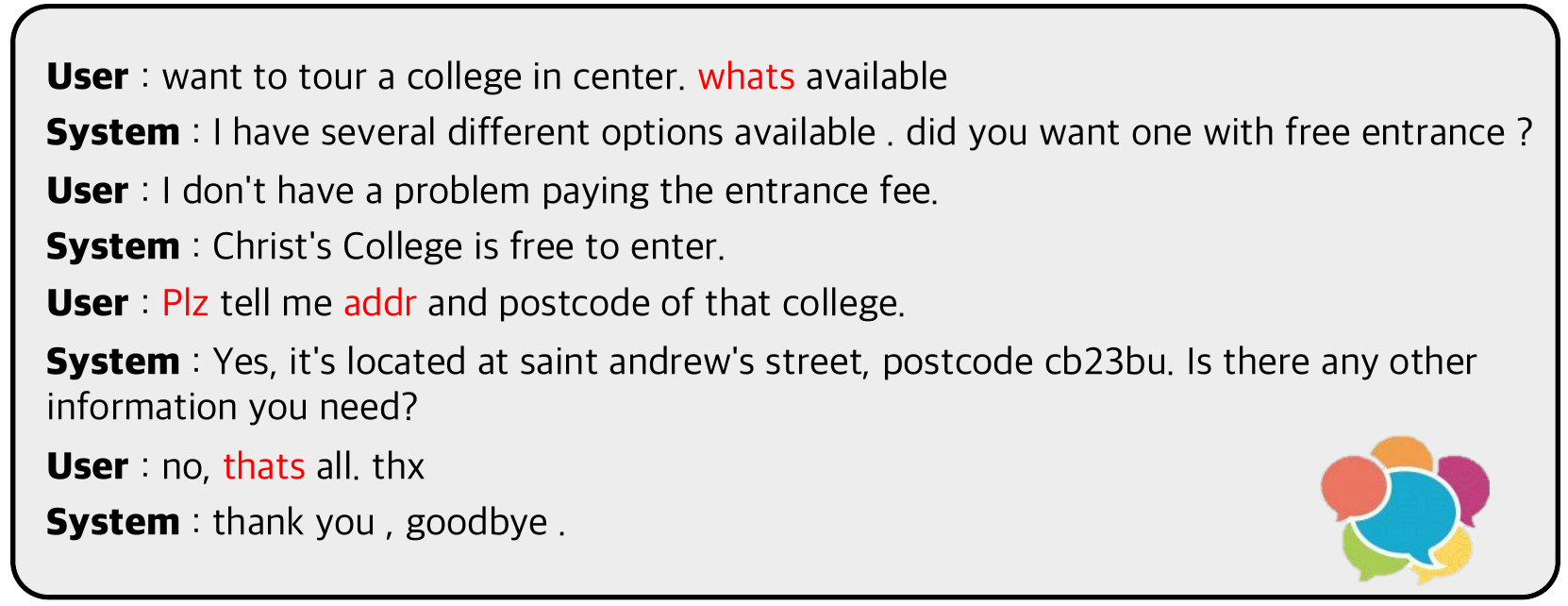} & 
    \hspace{-3mm}
    \includegraphics[height=3.10cm,width=7.80cm]{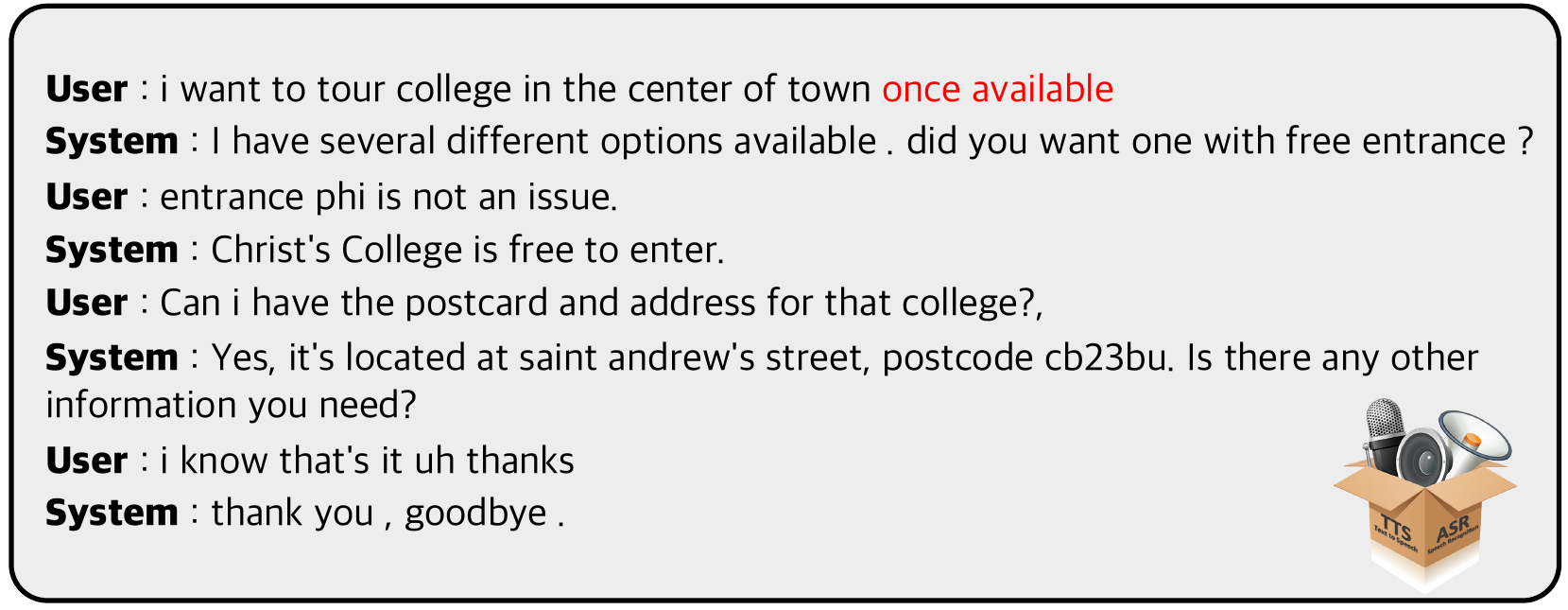} \\
    \vspace{3mm}
    (e) Typos \vspace{0mm} & 
    (f) Speech errors \hspace{-0mm} \\ 
    \hspace{-4mm}
    \includegraphics[height=3.15cm,width=7.80cm]{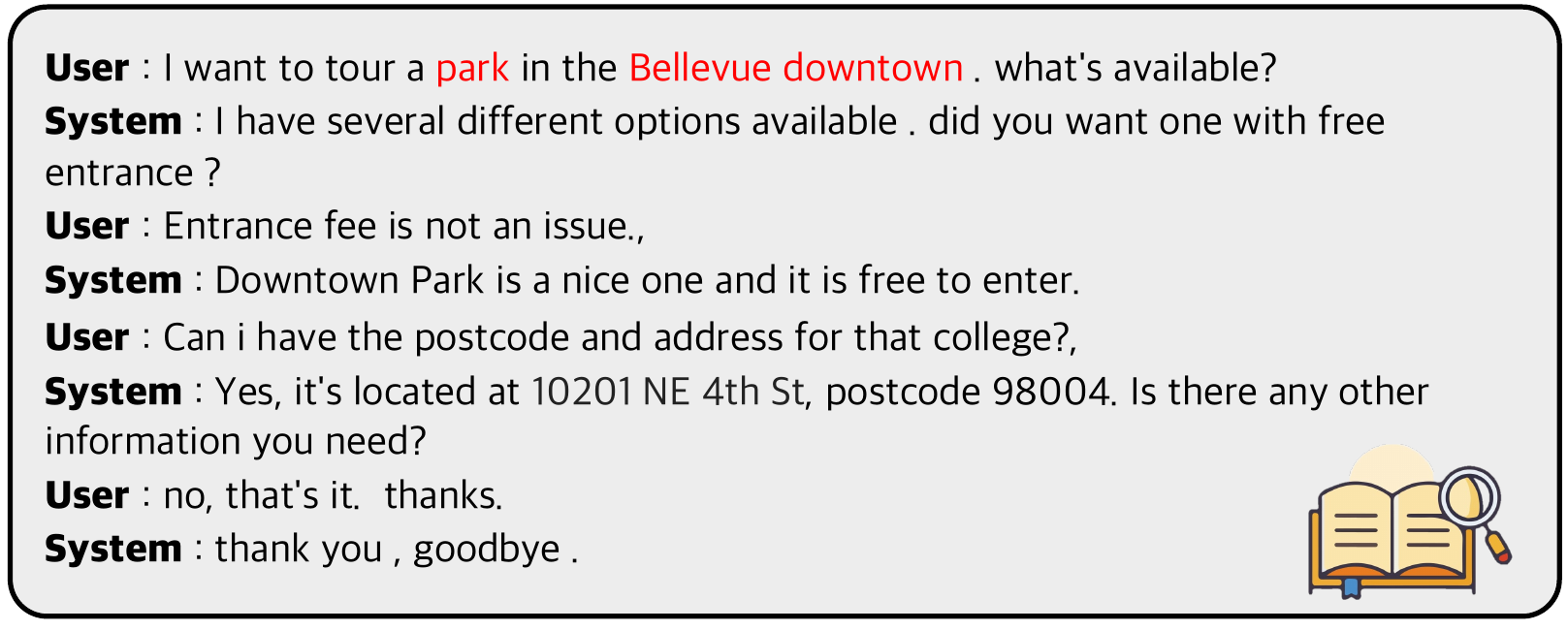} &
    \hspace{-3mm}
    \includegraphics[height=3.15cm,width=7.80cm]{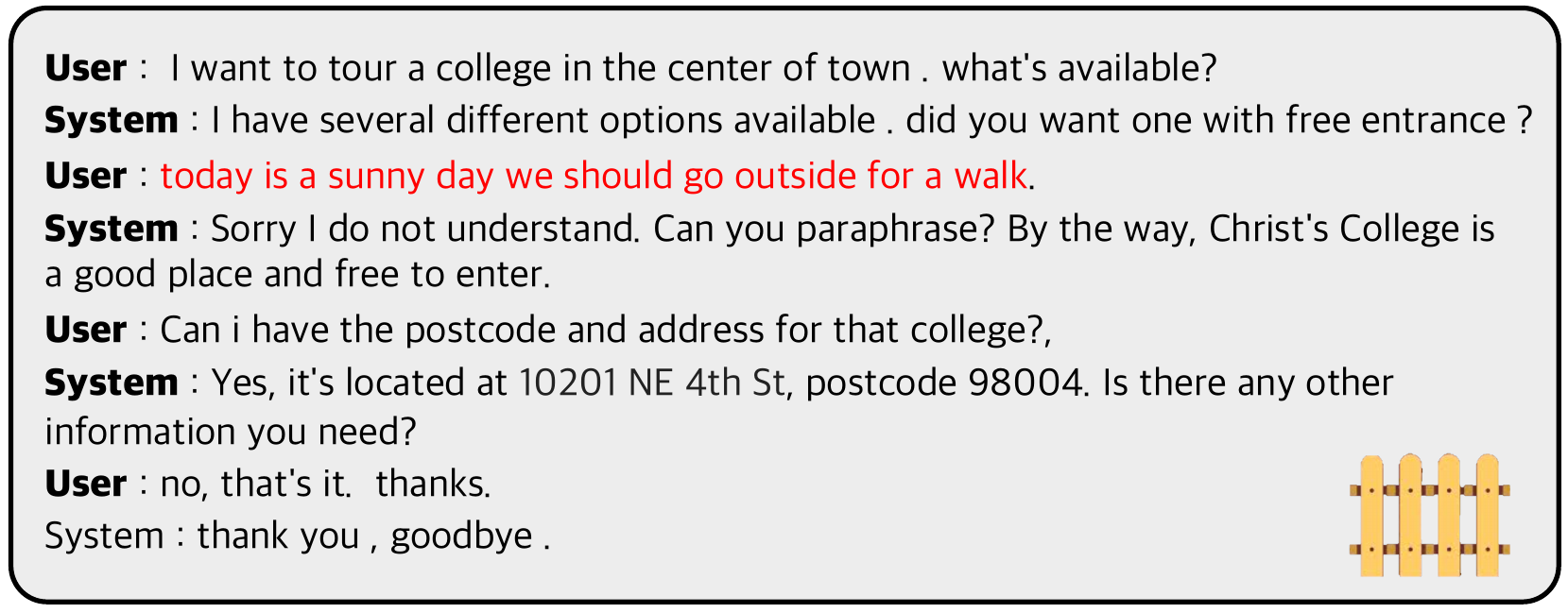} \\

    (g) Unseen entities \vspace{0mm} & 
    (h) Out-of-domain utterance \hspace{-0mm} \\    
  \end{tabular}
  \vspace{-0mm}
  \caption{Illustration of different language perturbations in the robustness diagnostic checklist. The standard dialog example is shown in (a). Based on it, (b)-(e) are four types of language variations~\iconvar{}, (f) shows speech error~\iconasr{}, (e) shows unseen entities~\iconent{}, and (h) shows out-of-domain utterance~\iconood{}. In each case, some representative examples are highlighted in red text.
   }
  \vspace{-0mm}
  \label{fig:robustness_examples}
\end{figure*}

\paragraph{Language Variations~\iconvar{}}
It is well-known that humans communicate using language with fairly large  variations such as different ways of expressions or personalized styles~\cite{sacks1978simplest}, while template-based crowd-sourcing fails in covering the linguistic variations~\cite{schegloff1977preference,moore2019conversational}. Specifically, we consider four types of variations in \shortname{}: 
$(\RN{1})$ {\em Paraphrase} widely exists among different users, who may present restatements of the meaning of a text or message using other words.
$(\RN{2})$ {\em Verbosity} describes a quality that users may express their intents using more words than  needed.
$(\RN{3})$ {\em Simplification} is a quality that users express their intents using fewer words to be concise.
$(\RN{4})$ {\em Typos} often result from illegitimate  abbreviations. In Figure~\ref{fig:robustness_examples}(b)-(e), we provide examples to illustrate these language variations.

\paragraph{Speech Errors~\iconasr{}} It is desirable that dialog systems can leverage automatic speech recognition (ASR) techniques to serve the speech modality, as in Amazon Alexa. However, almost all dialog systems have typically assumed that the user input is written text, and hoped that the system would seamlessly integrate with speech inputs. Recently, It has been empirically shown in~\citet{gopalakrishnan2020neural} that dialog systems trained on written data is very sensitive to various types of synthetic and actual ASR hypotheses in the dialog history. To bring attention to this gap, \shortname{} promotes speech robustness as an evaluation criterion. For example in Figure~\ref{fig:robustness_examples}(f), ``what's available'' can be transcribed as ``once available'' due to ASR deficiency, and a robust dialog system is expected to still correctly perceive user intents. 

\paragraph{Unseen Entities~\iconent{}}
Most existing DST methods are not designed to handle slot values that are not known to the tracker. The assumption that a pre-defined ontology exists for
the dialog and one can enumerate all possible values for each slot is often not valid in real-world scenarios. 
Even if such lists or dictionaries exist, they can be very large in size and highly dynamic~\cite{xu-hu-2018-end}. Therefore, {\em unseen entities} are common in dialogs, \ie  entities that are not observed during training, but appear in the testing stage. In Figure~\ref{fig:robustness_examples}(g), the entity \texttt{Bellevue downtown} is in the knowledge base but never appears in model training, a robust DST should be able to recognize it as a city/place, via generalizing from other similar entities learned during training.

\paragraph{Out-of-Domain Utterances~\iconood{}}
Most deployed task-oriented dialog systems are built for a closed set of target domains. Thus, they are fragile when dealing with out-of-domain (OOD) utterances~\cite{lee2019contextual}. Failure to detect OOD utterances often prevents the model from responding with an appropriate fallback action, hence leading to frustrating user experience. Therefore, it is important to endow task bots with the ability to detect OOD utterances for special handling~\cite{larson-etal-2019-evaluation}.  For example, in Figure~\ref{fig:robustness_examples}(h), the user suggests an excursion to a task bot trained in college consulting, which is out of the bot's scope. The bot is expected to raise a flag to label the utterance as an outlier, and guides the user to focus on the current domain. 

\paragraph{Collection Protocols}
The standard setting is sampled from MultiWOZ2.0 \cite{budzianowski2018multiwoz} but re-purposed in a few-shot learning setting. The language variations corpus is then created by workers on Amazon Mechanical Turks based on the standard corpus. To maximize the quality, we require workers in US locale and have a minimal previous approval rate of 90\%. Assignments are constructed at the turn level. Given a user utterance and associated dialog history, workers are required to answer four questions, what are the paraphrase, typos, verbose, and simplified versions of the user utterance. Moreover, in each assignment, the workers are instructed to exactly mention the slot values in the answers if the given user utterance has them. For the speech recognition errors setting, we employ the audio-level error simulation \cite{gopalakrishnan2020neural}, which generates audio signals from texts, adds noise into the audio, and then decodes the audio with an ASR model to obtain hypotheses. In particular, we employ Microsoft Cognition text-to-speech service to synthesize audio signals. After injecting background noise into the audio signals, we use the speech recognition service to obtain a corpus of Word Error Rate (WER) of 30\%. For the \texttt{reminder} domain that is applied for unseen entity evaluation, we firstly simulate several dialogs as seed scenarios using an agenda-based simulator and then randomly replace the slots in the dialogs with new values. Similar to constructing the language variations corpus, we then hire workers to rewrite the corpus as diverse and realistic as possible. Finally, the out-of-domain corpus is developed following \citet{lee2019contextual}. We randomly choose 50\% utterances in \texttt{DSTC}~\cite{henderson2014second} for the \texttt{Attraction} domain as the training set. 
For the test set, besides utterance from \texttt{DSTC}, we also introduce utterance from a diverse set of domains like \texttt{Stanford}~\cite{eric2017key}, \texttt{Reddit}, \texttt{Twitter}~\cite{sordoni2015neural} to evaluate the capability of handling different out-of-domain utterances.

\begin{table*}[t!]
    \centering
    \scriptsize
    \renewcommand{\arraystretch}{1.1}
    \setlength{\tabcolsep}{1.2mm}{
    \begin{tabular}{cccc|cccccccccccccc}
    \thickhline
    
& &
\multicolumn{2}{c|}{\texttt{Standard}} &
\multicolumn{2}{c}{\texttt{Paraphrase}} &
\multicolumn{2}{c}{\texttt{Simplification}} &
\multicolumn{2}{c}{\texttt{Typos}} &
\multicolumn{2}{c}{\texttt{Verbosity}} &
\multicolumn{2}{c}{\texttt{Speech ERR}} &
\multicolumn{2}{c}{\texttt{Unseen}} &
\multicolumn{2}{c}{\texttt{OOD}} \\

\cmidrule(l){3-4} \cmidrule(l){5-6} \cmidrule(l){7-8} \cmidrule(l){9-10} \cmidrule(l){11-12} \cmidrule(l){13-14} \cmidrule(l){15-16}  \cmidrule(l){17-18}

Model & Avg. & $\mathtt{JGA} \uparrow$ & $\mathtt{C} \uparrow$ &$\mathtt{JGA} \uparrow$ & $\mathtt{C} \uparrow$ &$\mathtt{JGA} \uparrow$ & $\mathtt{C} \uparrow$ &$\mathtt{JGA} \uparrow$ & $\mathtt{C} \uparrow$ &$\mathtt{JGA} \uparrow$ & $\mathtt{C} \uparrow$ &$\mathtt{JGA} \uparrow$ & $\mathtt{C} \uparrow$ &$\mathtt{JGA} \uparrow$ & $\mathtt{IC} \uparrow$ & $\mathtt{JGA} \uparrow$ & $\mathtt{F1} \uparrow$ \\
\hline
DAMD & - &
14.18 & 48.99 &
6.75 & 44.13 &
5.78 & 42.93  &
5.33& 42.58 &
7.08 & 42.56 &
9.1 & 45.94 &
- & - & - & -
\\
\gptft & 47.46 &
40.52 & 67.36 & 
31.36 & 62.72 &
28.82 & 59.44 &
22.31 & 54.15 &
30.40 & 54.16 &
31.41 & 65.95 &
28.28 & 51.29 & 
47.37 & 83.86
\\
\textsc{Soloist} & {\bf 59.09} &
{\bf 53.17} & {\bf 76.13} & 
{\bf 40.27} & {\bf 64.89} &
{\bf 37.18} & {\bf 63.61} &
{\bf 22.73} & {\bf 57.77} &
{\bf 38.21} & {\bf 65.71} &
{\bf 36.81} & {\bf 70.48} &
{\bf 69.05} & {\bf 96.98} & 
{\bf 56.28} & {\bf 96.18}
\\
\thickhline
    \end{tabular}
    }
    \caption{Overall results of baselines across all \shortname{} tasks. $\mathtt{C}$ indicates the $\mathtt{Combined}$ metric, $\mathtt{IC}$ denotes intent classification accuracy. Note that it is not straightforward to directly apply DAMD to Unseen and OOD tasks since it requires extra annotations. As such, we omit results of DAMD on these two tasks.}
    \label{table:overall}
\end{table*}

\section{Baselines}
For baselines, we consider three representative methods, holding state-of-the-art positions on existing benchmarks such as MultiWoZ~\cite{budzianowski2018multiwoz}.

\paragraph{DAMD}~\cite{zhang2019task} is a state-of-the-art modular system, where each dialog module is implemented using a neural network, and the whole system is trained in an end-to-end manner. 

\paragraph{GPT-2} represents a single multi-task learning model with impressive results on general language understanding and generation tasks. GPT-2 is an auto-regressive language model that leverages 12-24 layers of masked, multi-head self-attention Transformers. GPT-2 is pre-trained on extremely massive text data OpenWebText~\citep{gpt2}. It has demonstrated superior performance on characterizing human language data distribution and knowledge transfer. Given text prompts, GPT-2 can often generate fluent sentences. Its ancestral work GPT (with a smaller model size and less training data) has shown impressive results on language understanding tasks. In this paper, we consider \gptft{} as the approach of directly fine-tuning the pre-trained GPT-2 on a specific domain. Hence, \gptft{} can be viewed as \textsc{SOLOIST} without grounded pre-training, and serve as a strong baseline for both DST and E2E task. 

\paragraph{\textsc{Soloist} } represents recent model variants~\cite{Ham2020e2e,hosseini2020simple} to parameterize dialog system as a single auto-regressive model. \textsc{SOLOIST} subsumes different dialog modules (e.g. state tracker, dialog policy, response generator) into a single Transformer model. It has the similar capability with GPT-2 in understanding and generating natural language sentences but is pre-trained on large heterogeneous dialog corpora to gain additional capability of grounding text response in user goals and real-world knowledge for task completion \cite{peng2020soloist,gao2020robust}.


\paragraph{Training}~ We leverage the pre-trained checkpoints from the corresponding work, and fine-tune them on \shortname{}. Each domain is trained separately. We train our models with Adam with initial learning rate 5e-5 and batch size 1 for 20 epochs. 
To allow for fair comparisons with the both models, we do not tune hyper parameters or training settings for each model.

\paragraph{Evaluation.}
The \shortname{} benchmark follows the same evaluation model as GLUE~\cite{wang2018glue} or Kaggle\footnote{\url{https://www.kaggle.com/}}. To evaluate a system on the benchmark, one must run the system on the provided test data for the tasks, then upload the results to the website \url{http://aka.ms/raddle} for scoring 
The benchmark site shows per-task scores and a macro-average of those scores to determine a system’s position on the leaderboard. The website also provides fine- and coarse-grained results on the robustness diagnostic datasets. We will provide human evaluation services for top-ranked submissions on a bimonthly basis. The human evaluation protocol follows \citep{peng2020soloist, li2020results} 

\section{Benchmark Results}

\subsection{Overall Results}
We first present the results of baseline methods across all tasks  on the \shortname{} benchmark in Table \ref{table:overall}. As shown, \gptft{} fine-tuned with domain-specific dialog corpora outperforms the strong modular-based method DAMD. This highlights the efficacy of pre-trained language models. \textsc{Soloist} is the best-performing model and improves upon \gptft{} over 10 points in terms of average score, and consistently performs better than \gptft{} across all the tasks. These strong results indicate that large-scale task-specific pre-training on dialog corpora is crucial for effective and robust task adaptation.


\subsection{Robustness Diagnostic Checklist Results}

\begin{table*}[t!]
    \centering
    \scriptsize
    \renewcommand{\arraystretch}{0.9}
    \setlength{\tabcolsep}{1.0mm}{
    \begin{tabular}{ccccccccccccccccc}
    \toprule
    
\multicolumn{2}{c}{} &
\multicolumn{3}{c}{\texttt{Attraction}} &
\multicolumn{3}{c}{\texttt{Train}} &
\multicolumn{3}{c}{\texttt{Hotel}} &
\multicolumn{3}{c}{\texttt{Restaurant}} \\
\cmidrule(l){3-5} \cmidrule(l){6-8} \cmidrule(l){9-11} \cmidrule(l){12-14} \cmidrule(l){15-17} 

\multicolumn{2}{c}{{\it Task}} &$\mathtt{Inform} \uparrow$ & $\mathtt{Success} \uparrow$ & $\mathtt{BLEU} \uparrow$ & 
$\mathtt{Inform} \uparrow$ & $\mathtt{Success} \uparrow$ & $\mathtt{BLEU} \uparrow$ & 
$\mathtt{Inform} \uparrow$ & $\mathtt{Success} \uparrow$ & $\mathtt{BLEU} \uparrow$ & 
$\mathtt{Inform} \uparrow$ & $\mathtt{Success} \uparrow$ & $\mathtt{BLEU} \uparrow$ \\
\midrule
\multicolumn{2}{c}{{Env-0}} &
79.00 & 61.00 & 13.33 & 78.28 & 68.18 & 11.73 & 71.00 & 44.00 & 10.21 & 84.00 & 53.00 & 12.20 \\
\midrule
\multirow{4}{*}{\rotatebox{90}{Env-1}} &
{\it Para.} & 
71.00 & 51.00 & 12.40 & 81.31 & 71.72 & 11.74 & 66.50 & 35.50 & 9.40 & 70.50 & 40.00 & 12.09 \\
& {\it Simp.} & 
 63.00 & 47.00 & 12.40 & 78.28 & 69.19 & 11.97 & 57.50 & 32.50 & 9.45 & 68.00 & 43.00 & 12.23 \\
& {\it Typo} & 
 68.00 & 49.00 & 11.99 & 78.28 & 69.19 & 11.72 & 53.50 & 30.50 & 9.76 & 63.00 & 37.00 & 11.57 \\
& {\it Verbo.} & 
 76.00 & 54.00 & 12.79 & 75.25 & 67.17 & 11.97 & 61.50 & 40.00 & 10.41 & 71.00 & 43.50 & 11.43  \\ 
 \midrule

\multirow{4}{*}{\rotatebox{90}{Env-2}} &
{\it Para.} & 63.00 & 44.00 & 12.41 & 75.76 & 66.16 & 11.70 & 56.00 & 33.00 & 9.96 & 66.00 & 40.50 & 12.29 \\
& {\it Simp.} & 
 58.00 & 45.00 & 12.40 & 76.77 & 65.66 & 11.74 & 56.00 & 33.00 & 9.52 & 68.00 & 42.50 & 12.25 \\
& {\it Typo} & 
 60.00 & 41.00 & 11.75 & 75.25 & 66.67 & 11.67 & 49.00 & 27.50 & 10.08 & 52.50 & 30.50 & 11.62 \\
& {\it Verbo.} & 
 74.00 & 53.00 & 12.46 & 72.73 & 64.65 & 11.50 & 56.50 & 37.00 & 9.92 & 68.00 & 42.00 & 10.91  \\ 
 \midrule

\multirow{4}{*}{\rotatebox{90}{Env-3}} &
{\it Para.} & 
 63.00 & 39.00 & 12.48 & 78.28 & 64.65 & 11.26 & 59.00 & 35.00 & 10.08 & 63.50 & 34.00 & 11.67 \\
& {\it Simp.} & 
63.00 & 43.00 & 11.37 & 76.77 & 63.64 & 11.21 & 53.00 & 27.00 & 9.68 & 66.50 & 31.00 & 11.81 \\
& {\it Typo} & 
62.00 & 33.00 & 11.13 & 74.24 & 61.11 & 11.14 & 46.50 & 23.00 & 9.60 & 52.00 & 24.00 & 10.82 \\
& {\it Verbo.} & 
75.00 & 50.00 & 11.25 & 72.73 & 58.59 & 11.30 & 56.00 & 34.00 & 10.04 & 64.00 & 33.50 & 10.88  \\ 
\bottomrule 

    \end{tabular}
    }
    \caption{End-to-end Evaluation on \shortname{} and environments mixed with different ratios of user language variations. Env-X denotes replacing original test set with randomly sampled language variations examples, 10\%, 50\%, 80\%, respectively. }
    \label{table:e2e}
    \vspace{-2mm}
\end{table*}

\begin{figure*}[t!]
  \vspace{-0mm}\centering

    \includegraphics[width=2\columnwidth]{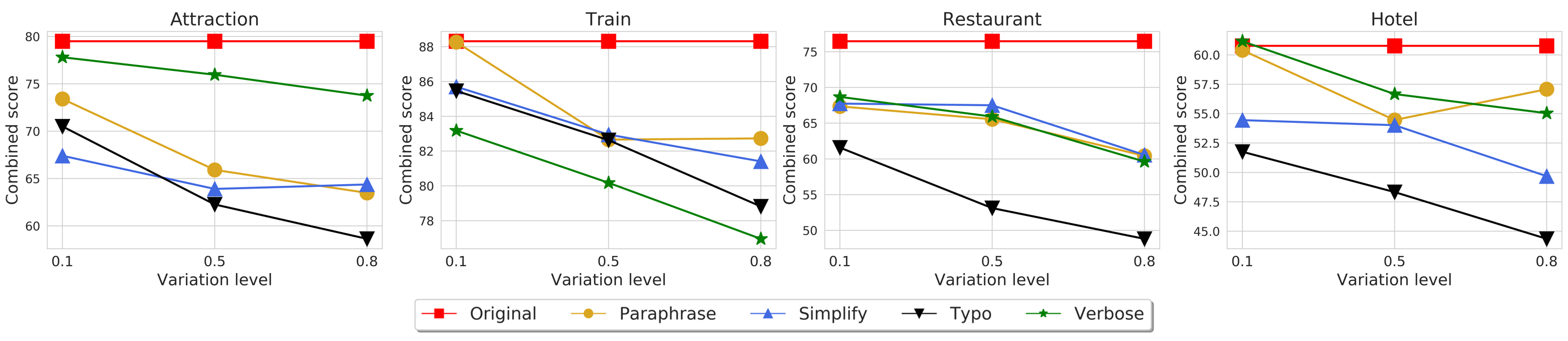}

  \caption{Evaluation results of \textsc{Soloist} on different levels of language variation corpus.}
  \vspace{-2mm}
  \label{fig:e2e_curve}
\end{figure*}

\begin{table}[htbp]
    \centering
    \scriptsize
    \renewcommand{\arraystretch}{0.9}
    \setlength{\tabcolsep}{1.0mm}{
    \begin{tabular}{cccccc}
    \toprule
    
\multicolumn{2}{c}{} &
\multicolumn{4}{c}{$\mathtt{Joint\ Goal\ Accuracy} \uparrow$}  \\

\cmidrule(l){3-6} 

\multicolumn{2}{c}{{\it Task}} &$\texttt{Attraction} $ & $\texttt{Train} $ & $\texttt{Hotel} $ & $\texttt{Restaurant}$  \\
\midrule
\multicolumn{2}{c}{{Env-0}} &
56.10 & 60.47 & 29.10 & 67.01 \\

\midrule
\multirow{4}{*}{\rotatebox{90}{Env-1}} &
{\it Para.} & 45.45 & 53.39 & 25.67 & 55.59 \\
& {\it Simp.} & 39.48 & 53.39 & 24.37 & 53.54 \\
& {\it Typo} & 34.55 & 36.97 & 16.96 & 42.69\\
& {\it Verbo.} & 45.45 & 47.79 & 23.82 & 55.82\\
 
 \midrule

\multirow{4}{*}{\rotatebox{90}{Env-2}} &
{\it Para.} & 43.38 & 46.61 & 22.24 & 50.91 \\
& {\it Simp.} & 37.66 & 42.18 & 22.98 & 50.46 \\
& {\it Typo} & 29.87 & 21.04 & 13.62 & 31.74 \\
& {\it Verbo.} & 43.90 & 39.04 & 21.59 & 54.11 \\
 \midrule

\multirow{4}{*}{\rotatebox{90}{Env-3}} &
{\it Para.} & 42.60 & 45.03 & 22.06 & 51.03 \\
& {\it Simp.} & 37.92 & 39.92 & 21.59 & 48.97 \\
& {\it Typo} & 29.09 & 19.37 & 12.60 & 30.25 \\
& {\it Verbo.} & 43.90 & 36.18 & 20.67 & 52.05 \\
 \bottomrule

    \end{tabular}
    }
    \caption{State tracking evaluation on \shortname{} and environments mixed with different ratios of language variations}
    \label{table:dst}
\end{table}

\begin{table*}[htbp]
    \centering
    \scriptsize
    \renewcommand{\arraystretch}{0.8}
    \setlength{\tabcolsep}{1.0mm}{
    \begin{tabular}{ccccccccccccccccc}
    \toprule
    
\multicolumn{2}{c}{} &
\multicolumn{3}{c}{\texttt{Attraction}} &
\multicolumn{3}{c}{\texttt{Train}} &
\multicolumn{3}{c}{\texttt{Hotel}} &
\multicolumn{3}{c}{\texttt{Restaurant}} \\
\cmidrule(l){3-5} \cmidrule(l){6-8} \cmidrule(l){9-11} \cmidrule(l){12-14} \cmidrule(l){15-17} 

\multicolumn{2}{c}{{\it Task}} &$\mathtt{Inform} \uparrow$ & $\mathtt{Success} \uparrow$ & $\mathtt{BLEU} \uparrow$ & 
$\mathtt{Inform} \uparrow$ & $\mathtt{Success} \uparrow$ & $\mathtt{BLEU} \uparrow$ & 
$\mathtt{Inform} \uparrow$ & $\mathtt{Success} \uparrow$ & $\mathtt{BLEU} \uparrow$ & 
$\mathtt{Inform} \uparrow$ & $\mathtt{Success} \uparrow$ & $\mathtt{BLEU} \uparrow$ \\
\midrule
\multicolumn{2}{c}{{SR-0}} &
79.00 & 61.00 & 13.33 & 78.28 & 68.18 & 11.73 & 71.00 & 44.00 & 10.21 & 84.00 & 53.00 & 12.20 \\
\midrule
\multicolumn{2}{c}{{SR-20}} &
72.00 & 53.00 & 13.37 & 77.78 & 67.17 & 12.37 & 62.00 & 39.00 & 9.99 & 73.50 & 44.50 & 11.38 \\
\multicolumn{2}{c}{{SR-30}} &
74.00 & 52.00 & 13.10 & 73.74 & 62.63 & 11.68 & 52.50 & 30.00 & 9.90 & 69.50 & 37.50 & 11.34 \\
\bottomrule

    \end{tabular}
    }
    \caption{Evaluation results of \textsc{Soloist} with different levels of speech errors. SR denotes speech errors. SR-X means corpus with X\% Word Error Rate (WER). }
    \label{table:wer}
\end{table*}

\begin{table*}[htbp]
    \centering
    \scriptsize
    \renewcommand{\arraystretch}{1.1}
    \setlength{\tabcolsep}{1.0mm}{
    \begin{tabular}{cccccccccccccccccccc}
    \toprule
    
\multicolumn{2}{c}{} &
\multicolumn{4}{c}{\texttt{DSTC}} &
\multicolumn{4}{c}{\texttt{Stanford}} &
\multicolumn{4}{c}{\texttt{Reddit}} &
\multicolumn{4}{c}{\texttt{Twitter}} \\
\cmidrule(l){3-6} \cmidrule(l){7-10} \cmidrule(l){11-14} \cmidrule(l){15-18}

\multicolumn{2}{c}{{\it Task}} &$\mathtt{P} \uparrow$ & $\mathtt{R} \uparrow$ & $\mathtt{F1} \uparrow$ & $\mathtt{DST} \uparrow$ & 
$\mathtt{P} \uparrow$ & $\mathtt{R} \uparrow$ & $\mathtt{F1} \uparrow$ & $\mathtt{DST} \uparrow$ & 
$\mathtt{P} \uparrow$ & $\mathtt{R} \uparrow$ & $\mathtt{F1} \uparrow$ & $\mathtt{DST} \uparrow$ & 
$\mathtt{P} \uparrow$ & $\mathtt{R} \uparrow$ & $\mathtt{F1} \uparrow$ & $\mathtt{DST} \uparrow$ \\
\midrule
\multirow{2}{*}{\rotatebox{90}{0\%}} &

\gptft{} & - & - & - & 46.75  & - & - & - & 46.75  & - & - & - & 46.75  & - & - & - & 46.75  \\
& \textsc{Soloist} & - & - & - & 56.10  & - & - & - & 56.10  & - & - & - & 56.10  & - & - & - & 56.10  
 \\
\midrule
\multirow{2}{*}{\rotatebox{90}{10\%}} &
 \gptft{} & 98.86 & 90.58 & 94.54 & 42.88 & 96.08 & 27.53 & 42.79 & 41.92 & 94.59 & 19.66 & 32.56 & 42.10 & 95.35 & 23.03 & 37.10 & 42.45  \\
& \textsc{Soloist} & 99.45 & 95.81 & 97.60 & 49.48 & 99.12 & 63.48 & 77.40 & 47.96 & 98.25 & 31.46 & 47.66 & 46.89 & 97.73 & 24.16 & 38.74 & 47.96  \\
 \midrule

\multirow{2}{*}{\rotatebox{90}{20\%}} &
\gptft{} & 100.00 & 98.42 & 99.21 & 45.66 & 100.00 & 76.40 & 86.62 & 44.58 & 100.00 & 48.88 & 65.66 & 44.40 & 100.00 & 57.30 & 72.86 & 45.12  \\
&\textsc{Soloist} & 97.94 &  99.48 & 98.70 & 49.48 & 97.65 & 93.26 & 95.40 & 49.56 & 96.75 & 66.85 & 79.07 & 50.62 & 96.99 & 72.47 & 82.96 & 52.04  \\
 \midrule

\multirow{2}{*}{\rotatebox{90}{50\%}} &
\gptft{} & 98.96 & 100.00 & 99.48 & 50.52 & 100.00 & 82.58 & 90.46 & 48.13 & 
100.00 & 52.81 & 69.12 & 48.85 & 100.00 & 61.80 & 76.39 & 49.02  \\
&\textsc{Soloist} & 99.47 & 100.00 & 99.74 & 55.21 & 100.00 & 97.75 & 98.86 & 52.93 & 100.00 & 83.71 & 91.13 & 53.82 & 100.00 & 90.45 & 94.99 & 54.71 \\
\midrule

    \end{tabular}
    }
    \caption{Results of out-of-domain detection using varying size of training examples on different target domains. N\% denotes injecting out-of-domain utterances N\% of the training set. }
    \label{table:ood}
\end{table*}

\begin{table}[t!]
    \centering
    \scriptsize
    
    \setlength{\tabcolsep}{1.0mm}{
    \begin{tabular}{lccccc}
    \toprule

\multirow{2}{*}{Model} & \multicolumn{3}{c}{$\mathtt{Slot\ Acc.}$ $\uparrow$} & \multirow{2}{*}{$\mathtt{JGA} \uparrow$} & \multirow{2}{*}{$\mathtt{IC} \uparrow$} \\
\cmidrule(l){2-4}
 &\texttt{Name}  & \texttt{Time} & \texttt{Day} \\
\midrule
\gptft{} & 84.47 & 65.01 & 26.85 & 28.28 & 51.29  \\  
\textsc{Soloist} & {\bf91.00} & {\bf 90.89} & {\bf 78.21} & {\bf 69.05} & {\bf 96.98}  \\ 

    \bottomrule
    \end{tabular}
    }
    \caption{ Evaluation results on unseen entities.}
    \label{table:useen_entity}
    \vspace{-0mm}
\end{table}

Table \ref{table:overall} shows the overall performance of DST and E2E modeling under different variation settings. 

\paragraph{Language variations} It is noticeable that all the models incur significant performance drops under each type of variation. Among all variation types, \texttt{Typos} has the most substantial impact on both JGA and $\mathtt{Combined}$ score resulting in 10 to 20 points of drop in performance. This is expected as misspelled keywords pose significant challenges for state tracking. The influence of other three types of variations are also prominent. The results reveal that existing SOTA dialog models trained on limited task-specific examples are not robust enough to handle various types of user utterances.

\paragraph{Speech errors} We observe a clear degradation in all metrics for all models. This shows that during inference, models trained on textual data are sensitive and not robust to actual ASR hypotheses introduced in dialog history. 

\paragraph{Unseen entities} Without task-specific pre-training, \gptft{} only achieves less than 30\% of JGA and 51.20 of dialog act accuracy even on a simple domain with most of the common entity values. \textsc{Soloist} performs significantly better than \gptft{} by achieving 69.05\% JGA and 96.98 dialog act accuracy but remains imperfect. These results imply that task-specific pre-training can improve the generalization capability of models but is still far from enough for production environments. 

\paragraph{Out-of-domain utterances} It is non-trivial for conventional modular-based dialog systems to handle out-of-domain detection. It often requires an additional component to classify whether a user utterance as in-domain or not. As such, we omit the result of DAMD in our experiments. We observe that pre-trained models handle out-of-domain detection relatively well. \gptft{} achieves 83.96 F1 score while \textsc{Soloist} has 96.18 F1 score, which shows that task-specific pre-training can improve robustness of models to out-of-domain utterances.

\subsection{Robustness detailed case studies}

First, to better understand the impact of different language variations, we evaluated \textsc{Soloist} on the corpus of different variation levels. 
Env-0 denotes the standard corpus, while Env-1, Env-2, Env-3 represent that 10\%, 50\% 80\%  of the standard corpus is replaced with language variation examples, respectively.  Table \ref{table:e2e} lists the detailed results on the end-to-end task and Table \ref{table:dst} shows the performance of state tracking. In general, the performance drops as the variation level increases for all types of variations across four domains. Even for a small variation level Env-1 (10\%), the performance drops significantly. We found that the degradation is mainly due to incorrectly tracked dialog states. Moreover, as depicted in Fig. \ref{fig:e2e_curve} and shown in Table \ref{table:e2e}, although the combined score drops on Env-2 and Env-3, \textsc{Soloist} still maintains good BLEU scores. 
These observations indicate that policy and response generation of \textsc{Soloist} are relatively robust to language variations and the dialog state tracking capability is the major bottleneck towards robust dialog models. An intriguing possibility to improve robustness is to apply adversarial training \cite{liu2020adversarial} to task-specific pre-training.

Next, similar to the experiments on language variations, we evaluated \textsc{Soloist} on corpus with different levels speech errors. Results are shown in Table \ref{table:wer}. We observe that compared with language variations, speech errors have a smaller impact on the performance for \textsc{Soloist}. It is noteworthy that the evaluation corpus we choose has considerably higher word error rates, ranging from 10\% to 30\%,  than a modern speech recognizer which usually has single-digit word error rate in quiet environments. We speculate that pre-trained dialog models trained on textual data has the potential to be deployed to smart home devices like Amazon Alexa, Apple Homepod. However, it still has defects when used in noisy environments such as smart assistant in cars or outdoor usage. There is less work on jointly pre-training speech and text modalities in dialog community. We believe that adding the speech modality to dialog pre-training may enhance robustness to speech errors.

Evaluation results on unseen entities are listed in Table \ref{table:useen_entity}. We observe that \gptft{} is able to handle unseen entities like \texttt{Name} and \texttt{Time} to some extent in this controlled experiment but fails in tracking \texttt{Day} properly, leading to inferior results in terms of joint goal accuracy and action selection. In contrast, with task-specific pre-training, \textsc{Soloist} substantially improves the performance in all the metrics. Nevertheless, for this simple task, string matching can effortlessly achieve near 100\% accuracy. Therefore, 69.05 joint goal accuracy is insufficient to affirm that \textsc{Soloist} is robust to unseen entities.
Incorporating knowledge into pre-training can be a solid basis for further research to improve robustness to unseen entities.

We also present the evaluation results of out-of-domain detection using varying sizes of training examples on different target domains in Table \ref{table:ood}. In the homologous DSTC domain~\cite{lee2019contextual}, \gptft{} performs similarly with \textsc{Soloist}. They are both able to identify out-of-domain utterances with near 100\% F1 score when injecting 50\% OOD data. However, \textsc{Soloist} leads 3 points in F1 score when trained using only 10\% data. In the other heterogeneous domains, \textsc{Soloist} performs consistently better than \gptft{}. In Reddit and Twitter domains that are distinct from DSTC, \textsc{Soloist} outperforms by over 20 points in F1 score when trained using 50\% data, showing that \textsc{Soloist} is more robust than \gptft{} to out-of-domain utterances. An inspiring observation is that injecting out-of-domain data can increase the performance of state tracking. While task specific pre-training helps with OOD detection, involving open-domain data into pre-training or initializing from open-domain dialog models such as DialoGPT \cite{zhang2019dialogpt} might further enhance robustness of dialog models.


\begin{figure}[t!]
  \vspace{-0mm}\centering
  \begin{tabular}{c c}
    &
    \hspace{-56mm}
    \includegraphics[height=0.34cm]{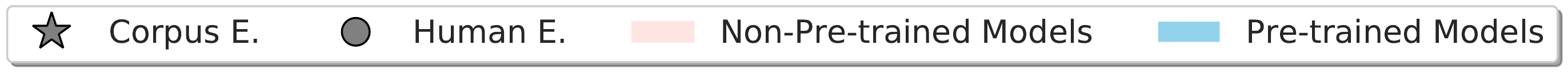}
     \hspace{-20mm}
     \\
    \hspace{-5mm}
    \includegraphics[height=3.6cm]{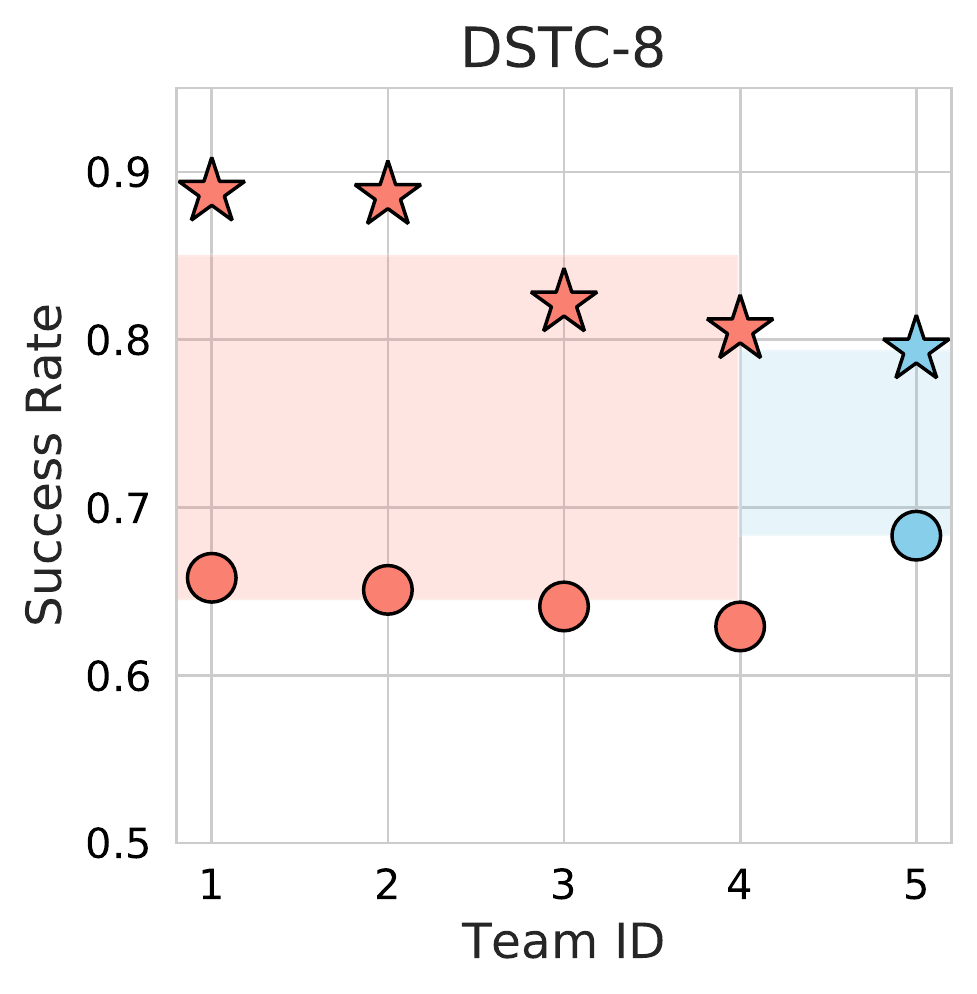}
    &
    \includegraphics[height=3.6cm]{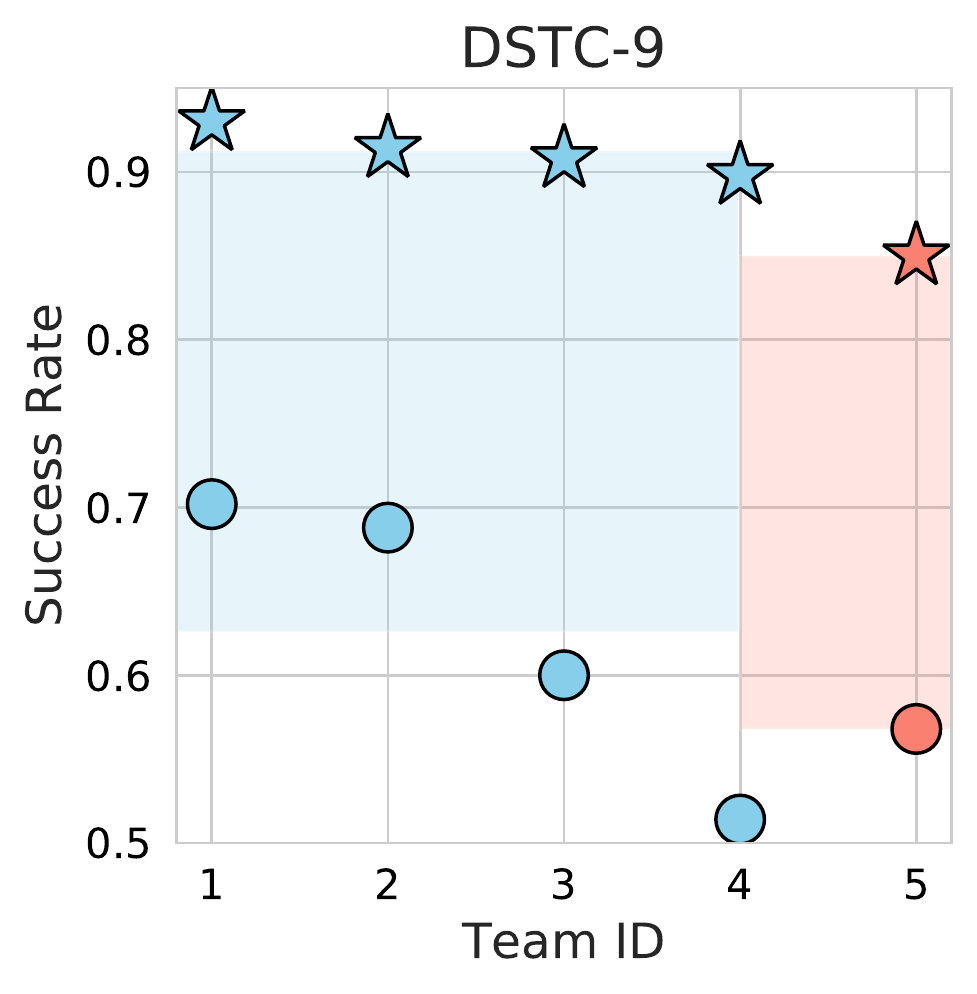} \\
    (a) DSTC-8 \vspace{2mm} 
    &
    (b) DSTC-9 \hspace{-0mm}  \\
  \end{tabular}
  \vspace{-2mm}
  \caption{Corpus and human evaluation for different models in two recent Multi-domain Dialog Challenges: (a) DSTC8 and (b) DSTC9. The regions indicate the gap between human and corpus evaluations for different types of models. We observe that 
  $(\RN{1})$ In DSTC8, Team 5 is the winner, and the only submission adopting pre-trained GPT-2 models; The performance discrepancy between the corpus and human evaluation is significantly smaller than other teams using modular-based methods without pre-training.
  $(\RN{2})$ a general trend shifting from modular based systems to pre-trained end-to-end systems. $(\RN{3})$  a substantial drop in performance which indicates that pre-trained methods remain sensitive to noisy inputs.
   }
  \vspace{-0mm}
  \label{fig:dstc}
\end{figure}

Finally, it is worth pointing out some important trends in the dialog research community, based on the DSTC challenge ~\cite{kim2019eighth,gunasekara2020overview} in the last 2 years (Figure~\ref{fig:dstc}). In DSTC8 ~\cite{kim2019eighth}, the winning submission by Team 5 is the only one that uses pre-trained models (GPT-2). When moving from corpus evaluation to human evaluation, it exhibits the least performance drop relative to other submissions, which is strong evidence to demonstrate robustness of pre-trained models. By the time of DSTC9~\cite{gunasekara2020overview}, the community have witnessed a general trend shift from modular systems to pre-trained end-to-end architectures. However, the significant performance gap between corpus evaluation and human evaluation indicates that pre-trained methods remain sensitive to noisy inputs. Such observations underscore the importance of robustness-oriented design and evaluation, for which \shortname{} fills a major void.

\section{Conclusion}
We introduce \shortname{}, a platform and collection of resources for evaluating and analyzing task-oriented dialog systems. We confirm the utility of grounded pre-training and transfer learning methods in dialog systems: pre-training improves generalization in a limited data setting, but still leaves room for improvement.  When evaluating these models on our diagnostic dataset, we find that they fail (often spectacularly) on many robustness test cases, suggesting possible avenues for future work. In summary, the question of how to design unified, efficient, robust models remains largely unexplored, and we believe that \shortname{} can provide fertile soil for addressing this challenge.

\bibliography{acl2020}
\bibliographystyle{acl_natbib}

\end{document}